\documentclass[12pt,draftclsnofoot,onecolumn]{IEEEtran}
\usepackage{times}
\usepackage{amsfonts}
\usepackage{cite,graphicx,amsmath,amsthm}
\usepackage{subfigure}
\usepackage{fancyhdr}
\usepackage{dsfont}
\usepackage{array,color}
\usepackage{bm}
\usepackage{float}
\usepackage{algorithm}
\usepackage{algpseudocode}
\usepackage{multirow}
\usepackage{bbm}
\usepackage{graphicx}
\usepackage{float}
\usepackage{subfigure}
\usepackage{ragged2e}
\usepackage{booktabs}
\usepackage{bbding}
\usepackage{multirow}
\usepackage{makecell}
\usepackage{amsmath,amssymb,amsfonts}
\usepackage[]{caption2}
\usepackage[table,xcdraw]{xcolor}
\usepackage{cite}

\begin{document}

\title{Robotic Wireless Energy Transfer in Dynamic Environments: System Design and\\Experimental Validation}

\author{Shuai Wang, Ruihua Han, Yuncong Hong, Qi Hao, Miaowen Wen, Leila Musavian, Shahid Mumtaz, and Derrick Wing Kwan Ng

\thanks{
Shuai Wang is with the Shenzhen Institute of Advanced Technology.

Ruihua Han, Yuncong Hong, Qi Hao are with the Southern University of Science and Technology.

Miaowen Wen is with the South China University of Technology.

Leila Musavian is with the University of Essex.

Shahid Mumtaz is with the Instituto de Telecomunica\~{o}es.

Derrick~Wing~Kwan~Ng is with the University of New South Wales.

Corresponding Author: Qi Hao.}

}

\maketitle
\vspace{-0.5in}
\begin{abstract}
Wireless energy transfer (WET) is a ground-breaking technology for cutting the last wire between mobile sensors and power grids in smart cities.
Yet, WET only offers effective transmission of energy over a short distance.
Robotic WET is an emerging paradigm that mounts the energy transmitter on a mobile robot and navigates the robot through different regions in a large area to charge remote energy harvesters.
However, it is challenging to determine the robotic charging strategy in an unknown and dynamic environment due to the uncertainty of obstacles.
This paper proposes a hardware-in-the-loop joint optimization framework that offers three distinctive features: 1) efficient model updates and re-optimization based on the last-round experimental data; 2)
iterative refinement of the anchor list for adaptation to different environments; 3) verification of algorithms in a high-fidelity Gazebo simulator and a multi-robot testbed.
Experimental results show that the proposed framework significantly saves the WET mission completion time while satisfying energy harvesting and collision avoidance constraints.
\end{abstract}

\IEEEpeerreviewmaketitle

\section{Introduction}

Powering massive Internet-of-Things (IoT) devices is a fundamental issue to realize intelligent monitoring, detection, manufacturing, and control in future smart cities.
Yet, it has always been regarded as a great challenge due to the limited size, vast volume, and sporadic nature of the IoT devices.
Recently, wireless energy transfer (WET) has been considered as a viable solution that deploys energy harvesters (EHs) on IoT devices such that the received radio frequency (RF) signals can be converted into electrical energies \cite{wet1}.
The harvested energy can be used for subsequent uplink communication with the wireless powered communication technology \cite{wpc}.
Compared with conventional energy-harvesting technologies such as solar, thermal, vibration, and magnetic resonant coupling, the advantages of WET are three-fold: 1) it involves no wire, no contact, fewer batteries, and represents a controllable energy supply \cite{wet1}; 2) the same RF circuitry for wireless communications can be reutilized for WET \cite{wet2}; 3) RF signals facilitates one-to-many charging due to the broadcast nature of wireless medium \cite{wet2}.

Current WET products (e.g., Powercast) only support short-range energy transmissions.
Robotic WET \cite{robowet1,robowet2,robowet3,robowet4} emerges as a promising solution, which mounts the energy transmitter on a mobile robot and navigates the robot through different regions in a large area so as to approach different EHs at different time slots.
Compared with unmanned aerial vehicle (UAV) WET \cite{uav1,uav2}, ground robots do not consume any propulsion energy to maintain stable hovering.
For instance, the motion power of a Turtlebot is 9.3\,Watts (i.e., 7 hours of operation time with a 14.8\,V 4400\,mAh battery), while the propulsion power of an UAV is above 100\,Watts \cite{uav1,uav2}.
However, ground robots need to face the complex collision avoidance problem on the ground as illustrated in Fig.~1.
Therefore, the UAV WET is suitable for time-sensitive applications, while the robotic WET is suitable for energy-sensitive applications.
Existing algorithms for robotic WET can be categorized into two types: global planning algorithms \cite{robowet1,robowet2,robowet3} and local planning algorithms \cite{local1,local2,local3}.
Global planning algorithms \cite{robowet1,robowet2,robowet3,robowet4} determine the anchor points, routes, and resources (e.g., charging time and beam directions at each anchor point), while local planning algorithms \cite{local1,local2,local3} periodically adjust the route for collision avoidance in a dynamic environment.
These global and local planning algorithms have been studied separately for robotic WET systems.
Therefore, it is necessary to integrate both for more efficient robotic charging, i.e., achieving smaller mission completion time while satisfying the energy harvesting and collision avoidance constraints.

Generally, it is challenging to determine the robotic charging strategy in an unknown and dynamic environment due to the uncertainty of obstacles.
Firstly, effective global and local planning algorithms are based on accurate mathematical models (e.g., robot motion time model).
Yet, parameters (e.g., distance matrix) in these models could be inaccurate in dynamic environments.
Secondly, joint optimization of anchors, routes, and resources is needed to adapt the planned route to different environments, which can be computationally expensive.
Finally, simulators should support close-to-reality features to verify the robustness of algorithms against practical uncertainties.

This paper provides three main contributions to address the above challenges.
Specifically, a hardware-in-the-loop (HIL) robotic WET system design is proposed, which allows efficient model updates and re-optimization based on the last-round experimental data.
Furthermore, a joint optimization framework based on the K-Chebychev density based spatial clustering of applications with noise (DBSCAN) and the successive local search is proposed to adjust the anchor list in different environments.
Finally, all the algorithms are implemented and tested in a high-fidelity Gazebo simulator and a multi-robot testbed.
The simulator and the testbed together form a digital-twin cyber-physical platform.
Extensive results based on the digital-twin platform are provided to verify the effectiveness of the proposed HIL joint optimization framework.
In the following, we first review the conventional robotic WET technologies before introducing our designs and implementations.

\begin{figure*}[!t]
 \centering
\subfigure[]{\includegraphics[height=0.4\textwidth]{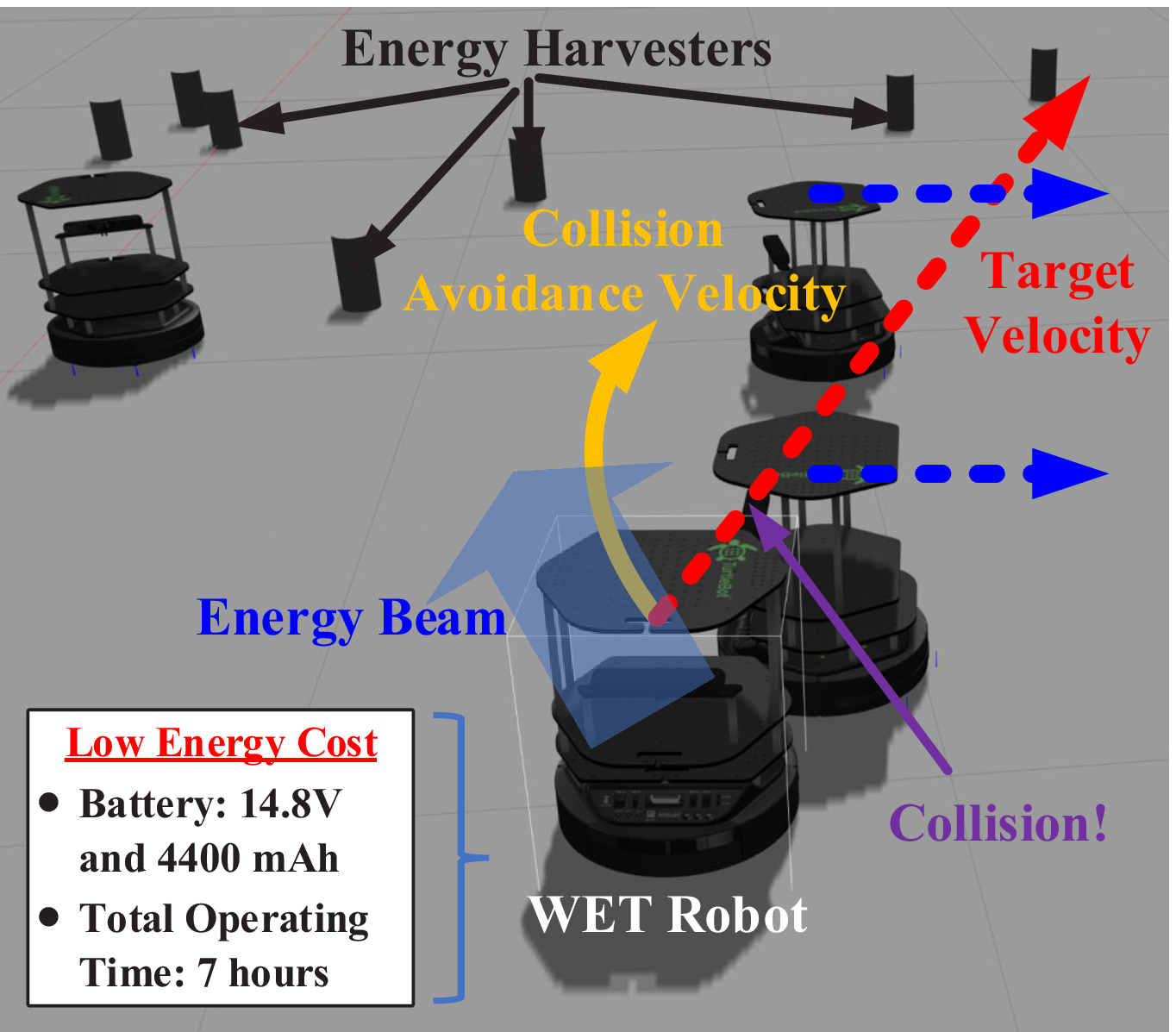}}
\subfigure[]{\includegraphics[height=0.4\textwidth]{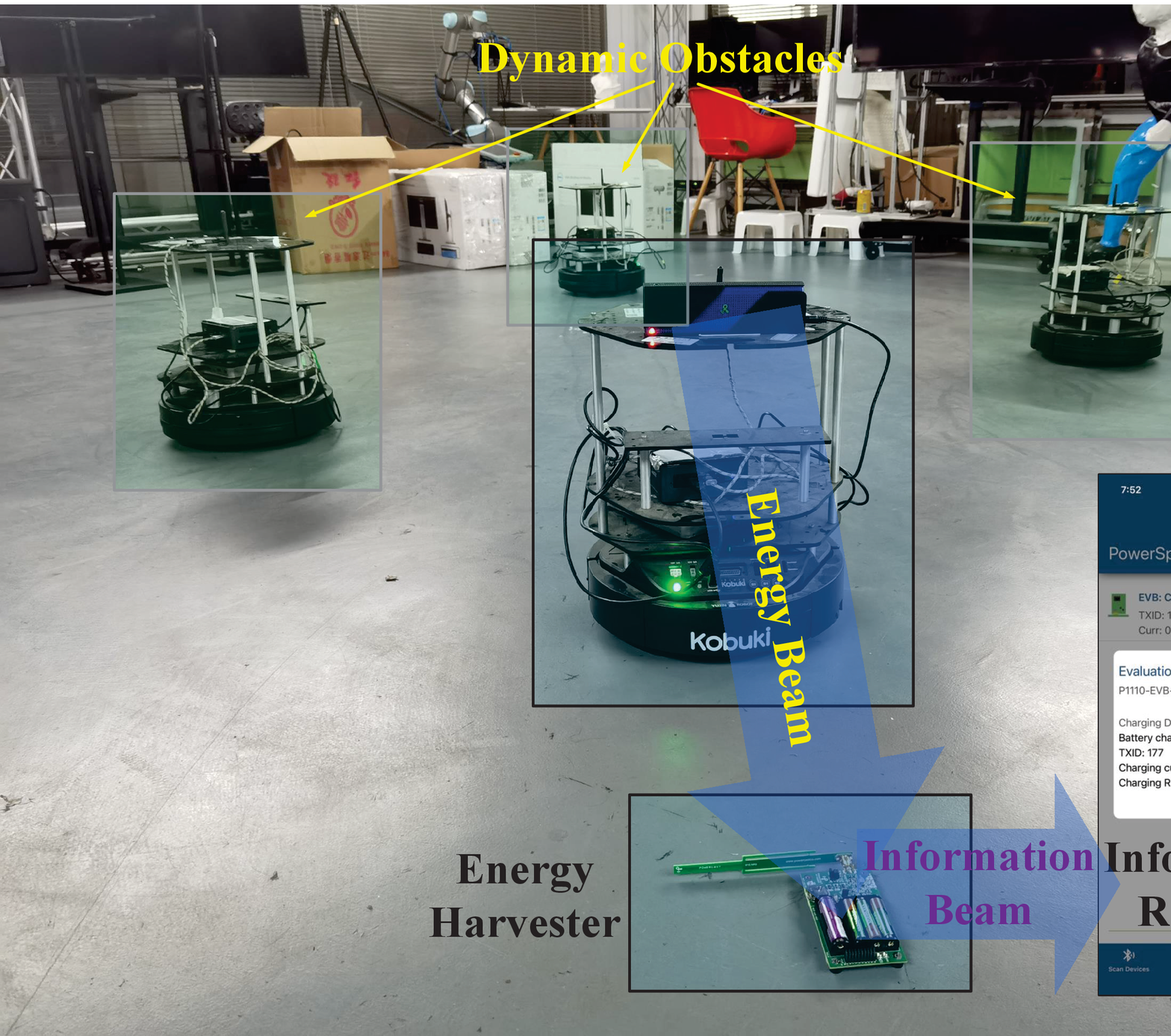}}
  \caption{A robotic WET system with dynamic obstacles: a) Gazebo simulation; b) multi-robot testbed.}
\end{figure*}

\section{Conventional Robotic Wireless Energy Transfer}

The robotic WET minimizes the total mission completion time (i.e., the sum of moving and charging time) by planning the anchor positions, routes, charging time and energy beams, while satisfying the energy harvesting requirements at all IoT devices \cite{robowet1,robowet2,robowet3}.
To achieve this goal, conventional robotic WET algorithms consist of 4 sequential steps:
\begin{itemize}
\item \textbf{Step 1}: Modeling.
Measurements are usually stored as look-up tables, which cannot be directly used for subsequent optimization.
We can transform the measurements into mathematical models and the parameters in these models are obtained by curve fitting.

\item \textbf{Step 2}: Anchor point generation.
This will divide the EHs into multiple clusters and assigns an anchor for each cluster where the robot will stop for a while to charge the surrounding EHs.

\item \textbf{Step 3}: Route planning.
The robot needs to visit the anchor points generated from Step 2.
Different routes (i.e., the sequence of anchor points) result in different motion time.

\item \textbf{Step 4}: Resource allocation.
The robot needs to determine the amount of time spent at each anchor point based on the models, anchors, and routes in Steps 1--3.
To fully exploit the degrees-of-freedom in the spatial domain, pencil-like energy-focusing beams can be shaped and steered towards the target devices.

\end{itemize}
In the following, we review the existing methods involved in each step.

\subsection{Mathematical Models}

To guarantee sufficient harvested energy at all nodes, the robot may visit-and-charge the IoT devices one-by-one.
Yet, this requires exceedingly long time spent in motion \cite{robowet1}.
A more time-efficient strategy is to let the robot simultaneously charge multiple devices at only a few positions.
These positions are called anchor points.
The robot motion time model measures the moving time from any anchor to another, which is used for subsequent anchor generation and route planning in Sections B and C.
This model is described by a directed graph, where the vertices represent the anchor points and the directed edges represent the feasible routes.
The route length is represented by a distance matrix, with the element at the $m$-th row and $j$-th column representing the distance from anchor $m$ to anchor $j$.
The motion time is calculated as distance over velocity.

To implement resource allocation in Section D, it is necessary to model the energy loss of robotic WET.
In general, wireless energies experience attenuations twice before being converted into electrical energies: 1) signal propagation in the air and 2) RF-to-current conversion in the EH.
\begin{itemize}
\item Wireless channel model is a function of the received RF power with respect to the transmitted RF power \cite{uav2}.
It can be categorized into statistical, deterministic, and quasi-deterministic methods.
Statistical models are low-complexity but inaccurate parameterized models.
Deterministic models (i.e., ray tracing) launch rays in various directions and traces the propagation, refection, scattering, and absorbtion using the electromagnetic field.
However, it requires exceedingly long modeling time.
Most industrial standards adopt quasi-deterministic channel models, which obtain hyper-parameters for different scenarios using ray tracing and generate wireless channels using some statistical method.

\item Energy harvesting model is a function of the harvested power with respect to the received RF power \cite{wet1}.
This function is nonlinear, since an energy harvesting circuit consists of nonlinear elements such as diodes.
According to \cite{nlmodel}, the energy harvesting model should satisfy the monotonicity, sensitivity, nonlinearity, and saturation properties.
Energy harvesting models can be categorized into piecewise linear model, logistic model, and sensitivity-based logistic model \cite{wet1}.
The former two models are low-complexity but only support partial properties.
The last model supports all properties but is a nonconvex function that may cause additional computational costs during the subsequent steps \cite{nlmodel}.
\end{itemize}
Details of the models can be found in Appendix A.

\subsection{Anchor Point Generation}

Anchor point generation can be viewed as a spatial clustering problem \cite{robowet3,cheby}.
In general, clustering for robotic WET can be categorized into distance-based and density-based methods.
In distance-based methods (e.g., k-means clustering), a distance metric is used to determine the similarity between EHs.
The method produces compact and spherical clusters around a set of centroids that are very sensitive to outliers.
On the other hand, density-based methods (e.g., DBSCAN) adopt a density threshold to distinguish the important EHs from the outliers.
As such, it can deal with unbalanced clusters and outliers pretty well.
Different from k-means clustering, DBSCAN generates arbitary shapes, which provide higher flexibility than k-means for anchor point generation.

\subsection{Route Planning}

With the positions of anchor points, the next step is to determine the visiting sequence of these points via route planning.
The route planning problem is a constrained discrete optimization problem \cite{robowet1,robowet2}, where the constraints guarantee: 1) the robot returns to the starting point; 2) the robot visits the selected vertices; 3) the planned path is connected.
Conventionally, tree search algorithms, e.g., branch-and-bound (B\&B), can be adopted for systematically pruning out ineffective solutions, leading to significant reduction of the computational complexity compared to exhaustive search while guaranteeing optimality.
However, its complexity is still high since the solution space grows exponentially with the number of anchor points.
Imitation learning emerges as a promising solution to solve the large-scale discrete optimization problem.
The core idea is to treat route planning as a classification problem and adopt a deep neural network to mimick the behavior of tree search algorithms.

\subsection{Resource Allocation}

When the WET robot moves along the optimized route, the remaining factors impacting the system performance are the resources allocated at each anchor point.
Common resources include the charging time (time-domain) and the energy beams (angle-domain) \cite{nlmodel}.
For charging time allocation, different anchor points along the route should be jointly considered.
This is because the charging time at the current location might have a long-term impact on future locations, as the harvested energy at IoT devices can be stored in batteries.
For beam allocation, there are two different types: multi-antenna based and directional-antenna based.
In particular, multi-antenna energy beamforming adjusts the power and phase at each antenna to form the desired beam directions.
In this case, the harvested energy is the multiplication of the charging time and the harvested power, which introduces non-convexity to the resource allocation problem.
Advanced optimization tools such as majorization minimization \cite{nlmodel} (which iterates between solving and finding a surrogate problem of the primal problem) can be used to obtain suboptimal solutions.
On the other hand, the directional-antenna beamforming has a sector shape and the robot needs to rotate itself to alter the beam direction.
Using the exhaustive search, the beam direction can be selected from a finite codebook with pre-designed beam patterns.

\section{Proposed Hardware-in-the-loop Joint Optimization Framework}

\begin{figure*}[!t]
\centering
\includegraphics[width=0.98\textwidth]{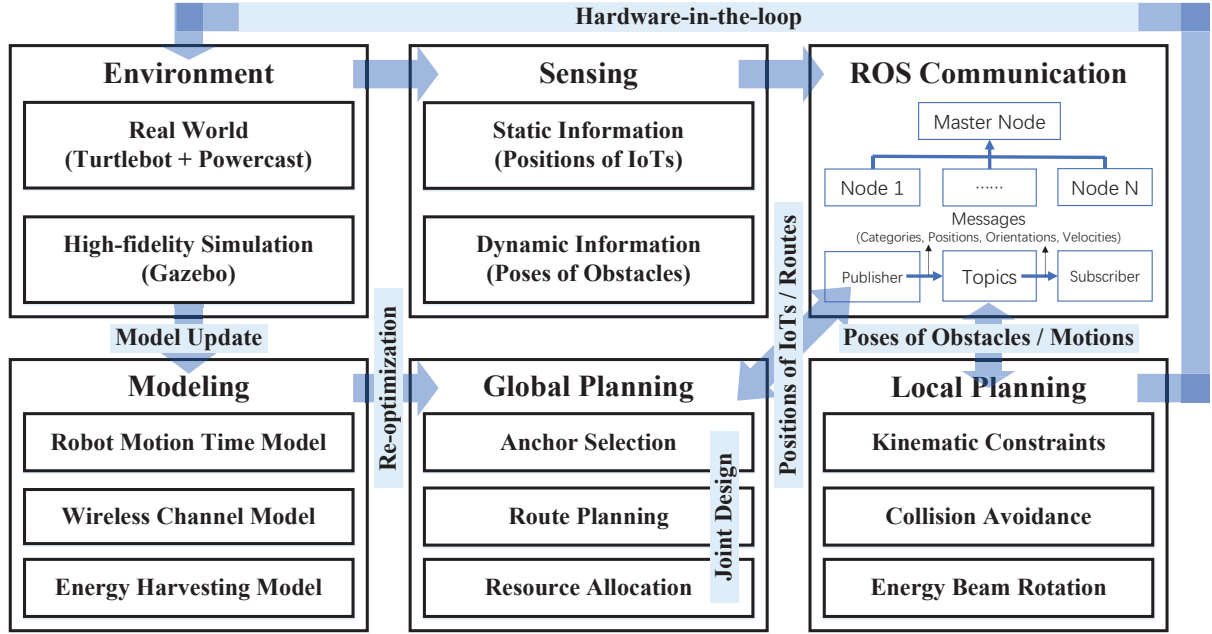}
\caption{Hardware-in-the-loop (HIL) joint optimization framework for robotic wireless energy transfer.}
\end{figure*}

\subsection{System Description}

Existing schemes assume that the WET robot has perfect knowledge of the environment, and the planned route can be directly adopted.
In practice, however, the WET robot needs to perform periodic sensing and avoid collisions with the surrounding obstacles.
This implies that the WET robot must adjust its route online, leading to a mismatch between the planned and actual routes.
To mitigate the mismatch, the HIL joint optimization framework is proposed in Fig.~2, which supports interactions among experiments, simulations, modeling, and planning based on the robot operating system (ROS) communication.
The goal is to allow efficient model updates and re-optimization based on the last-round experimental data.
For example, the initial motion time between two anchors is estimated as straight line distance over velocity.
But if the robot takes a detour due to the obstacles between the two anchors, the actual motion time would be longer than the estimated time.
With feedback of the actual motion time obtained from experiments, the motion time model becomes closer to reality, thereby improving the system performance.

Specifically, the system in Fig.~2 operates in an iterative manner.
Each iteration consists of the following operations.
\begin{itemize}
\item \textbf{Offline Planning Stage}. First, the models are fitted to the measurements obtained from the last-round experimental validation.
Measurements may include: 1) the amount of motion time when the robot travels from one anchor point to another; 2) the channel fading versus the charging distance; 3) the harvested power versus the incident power.
Then, the robot explores the environment for obtaining the current positions of EHs and building the static map based on simultaneous localization and mapping (SLAM).
Finally, with the fitted models and positions of EHs, the global planner generates the charging policy (including route, time, beam) offline.

\item \textbf{Online Validation Stage}.
First, the robot follows the target route produced in the offline stage.
Then, the robot performs object detection periodically using on-board sensors.
The sensor outputs (i.e., ego-positions, ego-velocities, and poses of other obstacles) are shared to the industrial computer via ROS communication for local planning.
Finally, the local planner adjusts the route constantly to avoid collisions.
This is achieved by computing a velocity vector that is closet to a target route provided by the global planner, while taking into account the robot kinematics and the chances of collisions.
The online stage is implemented in real environments or close-to-reality simulators, which would generate new measurements for next-round offline planning.
\end{itemize}
Below, we provide details of the new modules and features supported by the framework.

\subsection{Joint Optimization For Global Planning}

As shown in the lower-middle of Fig.~2, the proposed framework adopts a joint optimization framework for global planning, which has two major differences compared with conventional algorithms.
First, we adopt the K-Chebychev DBSCAN method instead of the k-means clustering for anchor point generation.
This is because k-means clustering can minimize the sum distance between the anchor point and its associated EHs.
However, for WET systems, the key factor affecting the system performance is the distance from each anchor to the farthest EH in its cluster \cite{cheby}.
Therefore, after executing the DBSCAN algorithm, we re-compute the anchors' positions as Chebyshev centers via the min-max optimization \cite{cheby}.
Furthermore, if the transmitter adopts a directional antenna, the coverage of an energy beam should also be considered, e.g., by adding an energy beamforming constraint \cite{wet1} to the min-max optimization problem \cite{cheby}.
Second, we add an anchor point selection module to prune out ineffective positions and navigate the robots away from dense traffic as possible.
This is realized via joint optimization based on the iterative local search framework \cite{robowet2} at the global planning layer.
The algorithm iterates between selecting a subset of anchor points among all the candidates and executing the route planning and resource allocation algorithms.
As such, the anchor positions are no longer fixed as in conventional robotic WET, but are adaptively optimized in different environments.
Details of the joint optimization can be found in Appendix B.

\subsection{Local Planning For Collision Avoidance}

Conventional robotic \cite{robowet1,robowet2,robowet3} or UAV-aided \cite{uav1,uav2} WET schemes ignore the local planning problem.
In practice, however, the robot must avoid collisions with any object residing in the environment, while making progress towards the next anchor point for wireless charging.
This is realized by the local planner shown in the lower-right of Fig.~2.
The local navigation algorithms depend on the mobility of obstacles.
If the observed obstacles are static, the key is to compute a velocity vector for the robot while taking into account the robot kinematics and dynamics.
If the obstacles are moving, the above approach can be applied by extrapolations of the observed velocities to estimate the future positions of obstacles.
The problem of collision avoidance becomes challenging when the obstacles are not simply moving with a constant speed, but are also intelligent decision-making agents that try to avoid collisions as well.
This is because each robot can only estimate the positions and velocities of other agents but cannot know their reactions and intents.
In this case, each robot needs to select a velocity outside the reciprocal velocity obstacle (RVO) region induced by other agents \cite{local3}.
It has been proved that the RVO provides a sufficient and necessary condition for a robot to avoid collisions with an obstacle moving with a known velocity but unknown intention \cite{local1}.
Note that the RVO scheme can be replaced by deep reinforcement learning (DRL), which can achieve a higher efficiency under a properly tuned reward function and a specific environment.
DRL adopts a policy network to map the environment state into the robot motions and is trained to converge to the actions with the maximum cumulative reward over several episodes.
However, the DRL-based approaches are computationally expensive and sensitive to sensor noises.

\subsection{Hardware-in-the-loop For Re-optimization}

The key feature of HIL is that algorithms are tested in close-to-reality simulators or real-world environments so as to make re-optimization possible \cite{hil}.
For simulation, we adopt Gazebo \cite{local1,local3}, a high-fidelity robotic evaluation platform that adopts Open Dynamics Engine for motion generation of the robots and Open Graphics Library Engine for the visualization of the world.
On top of the two engines, Gazebo realizes each physical object as a model that is composed of rigid bodies, joints, sensors, and interfaces for client programs to control the model.
In our experiment, the WET robot is modeled as a combination of various bodies, where different bodies are assigned different mass, friction, and bounce features.
The hinge joints among the bodies of a robot provide the physical mechanism to form kinematic and dynamic relationships such as rotations.
The EHs are modeled as cylinders.
The Gazebo implementation is shown in Fig.~1.

As for the experiment, we adopt the Turtlebot2 platform, equipped with an ultra-wide-band (UWB) sensor that produces the ego-location of the robot.
The onboard industrial computer controls the chassis as well as the direction of Powercast transmitter TX91503.
A mobile battery powers the chassis, sensor, computer, and the transmitter.
Communication among these devices is implemented via ROS, which is a distributed communication framework supporting integrative and heterogenous systems \cite{local3}.
In ROS, all processes that perform computations are implemented as nodes, where a master node controls the global system and slave nodes manage programs on each device.
ROS offers a message passing interface that provides inter-process communication among these nodes via topics.
A node sends a message by publishing it to a given topic and any node interested in the message can subscribe to the associated topic.

\section{Experimental Validation}

We consider the task of 1 WET robot charging 20 EHs.
The transmit power of Powercast TX91501 is 3\,Watts at 915\,MHz.
The transmit antenna is directional and its beamwidth is 130 degree.
The codebook contains 3 beam patterns, i.e., -65 to 65 degree, 55 to 185 degree, and 175 to 305 degree.
The receiver gain at Powercast P2110 is 6\,dBi.
The maximum velocity of WET robot is 0.2\,m/s.
The minimum energy to be harvested at each IoT device is 20\,mJ.
The pathloss at 1\,m is -31.6284\,dB and the pathloss exponent is 1.73 according the In-H channel model specified in 3GPP TR 38.901.
The sensitivity-based energy harvesting model in \cite{nlmodel} is adopted, with the parameters specified in Fig.~2 of \cite{nlmodel}.

First, the WET robot finds all the EHs via SLAM as shown in Fig.~3a.
Then Fig.~3b compares the results of joint optimization and sequential optimization in the perfect case.
It can be seen that the anchor points (i.e., blue squares) generated by joint optimization are slightly shifted from the centroids (i.e., red crosses) in order to facilitate the beam design.
Moreover, joint optimization only selects a subset of anchors from the candidates, allowing the robot to charge EHs in other clusters.
This is in contrast to the conventional sequential method, where the robot needs to visit all the anchors, leading to excessive motion time.
The above results corroborate discussions in Section III-B.

\begin{figure*}[!t]
 \centering
\subfigure[]{\includegraphics[height=0.48\textwidth]{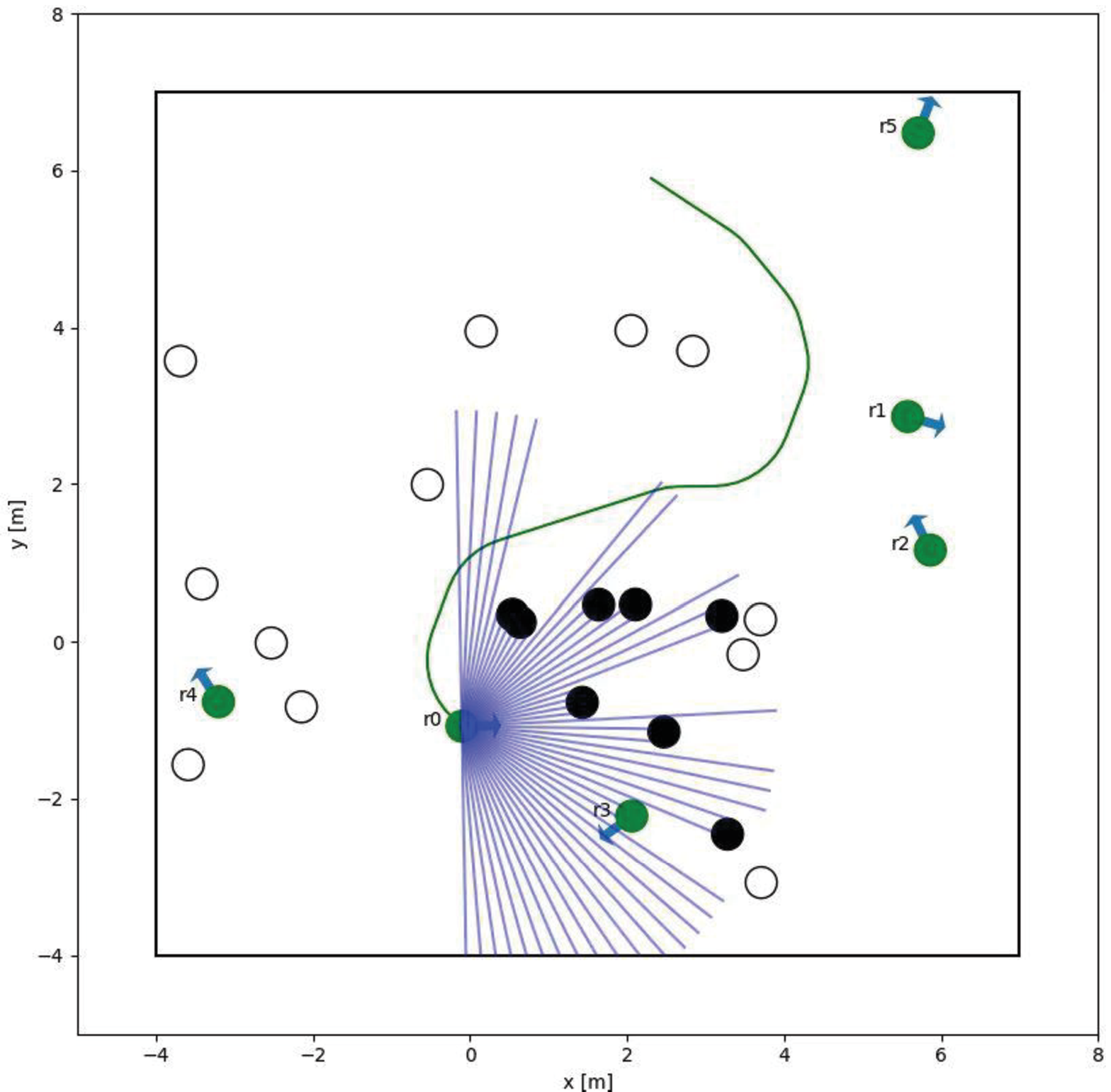}}
\subfigure[]{\includegraphics[height=0.48\textwidth]{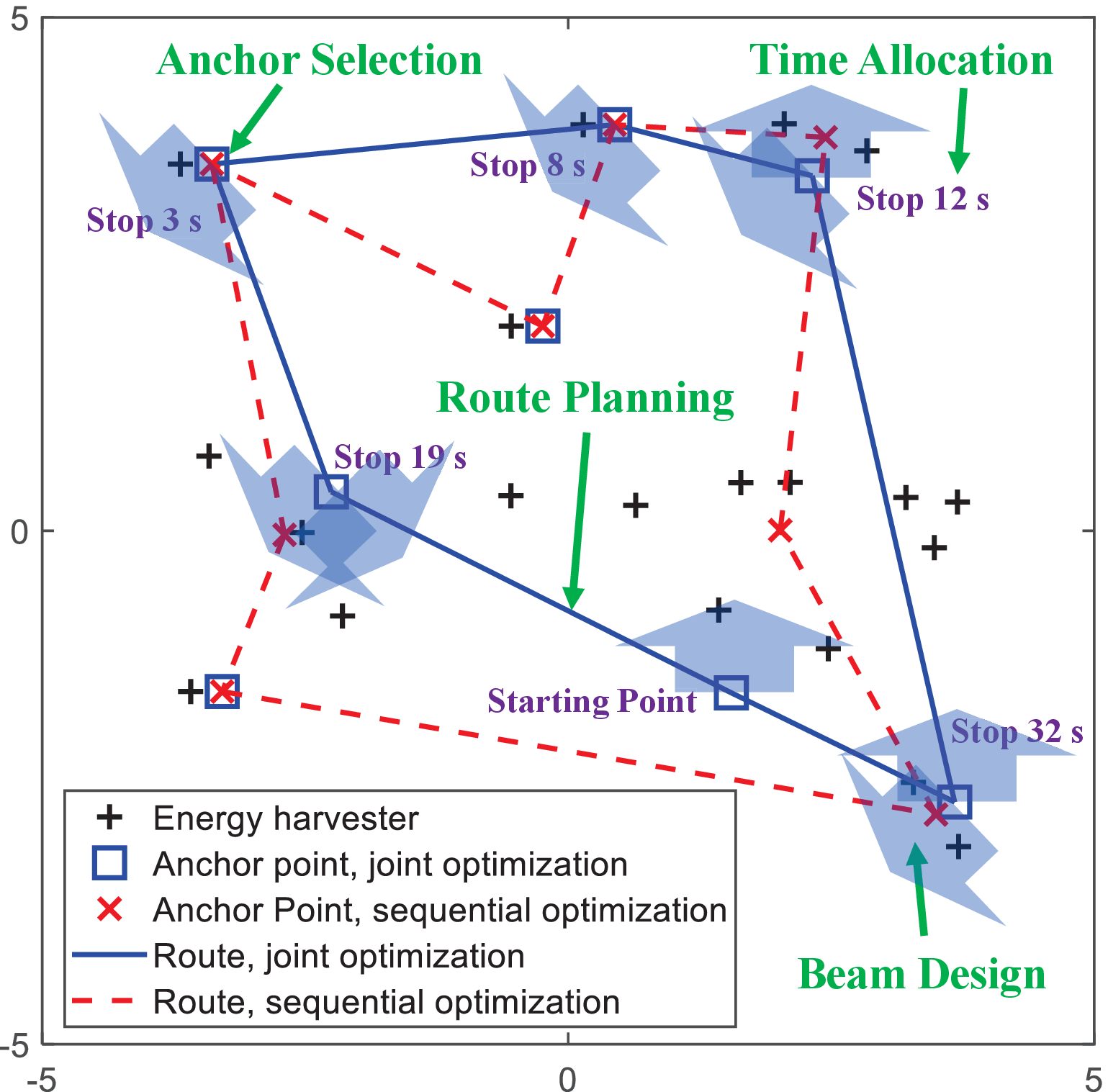}}
  \caption{a) The robot finds the EHs via SLAM; b) Comparison between the proposed joint optimization and conventional sequential optimization algorithms.}
\end{figure*}

The task in Fig.~3 is implemented in Gazebo, with 5 noncooperative robots moving randomly in the same environment.
The associated result is shown in Fig.~4a.
It can be seen that the blue route in Fig.~3b and the purple route in Fig.~4a share a high similarity.
However, in Fig.~4a, the WET robot takes a detour when returning to the starting point while the robot takes a straight way in Fig.~3b.
This is because the robot comes across other robots in the way in Gazebo and needs to find a tradeoff between avoiding collision and reaching the next anchor point.
Moreover, the robot spends additional time to rotate itself in Gazebo.
Consequently, the motion time in Gazebo is 147\,s, which is 50\% longer than that in the perfect case (i.e., 110.369\,s).
The motion time in the real world experiment is 150\,s, meaning that the Gazebo simulator and the real-world testbed can together form a digital-twin platform.

\begin{figure*}[!t]
 \centering
\subfigure[]{\includegraphics[width=0.6\textwidth]{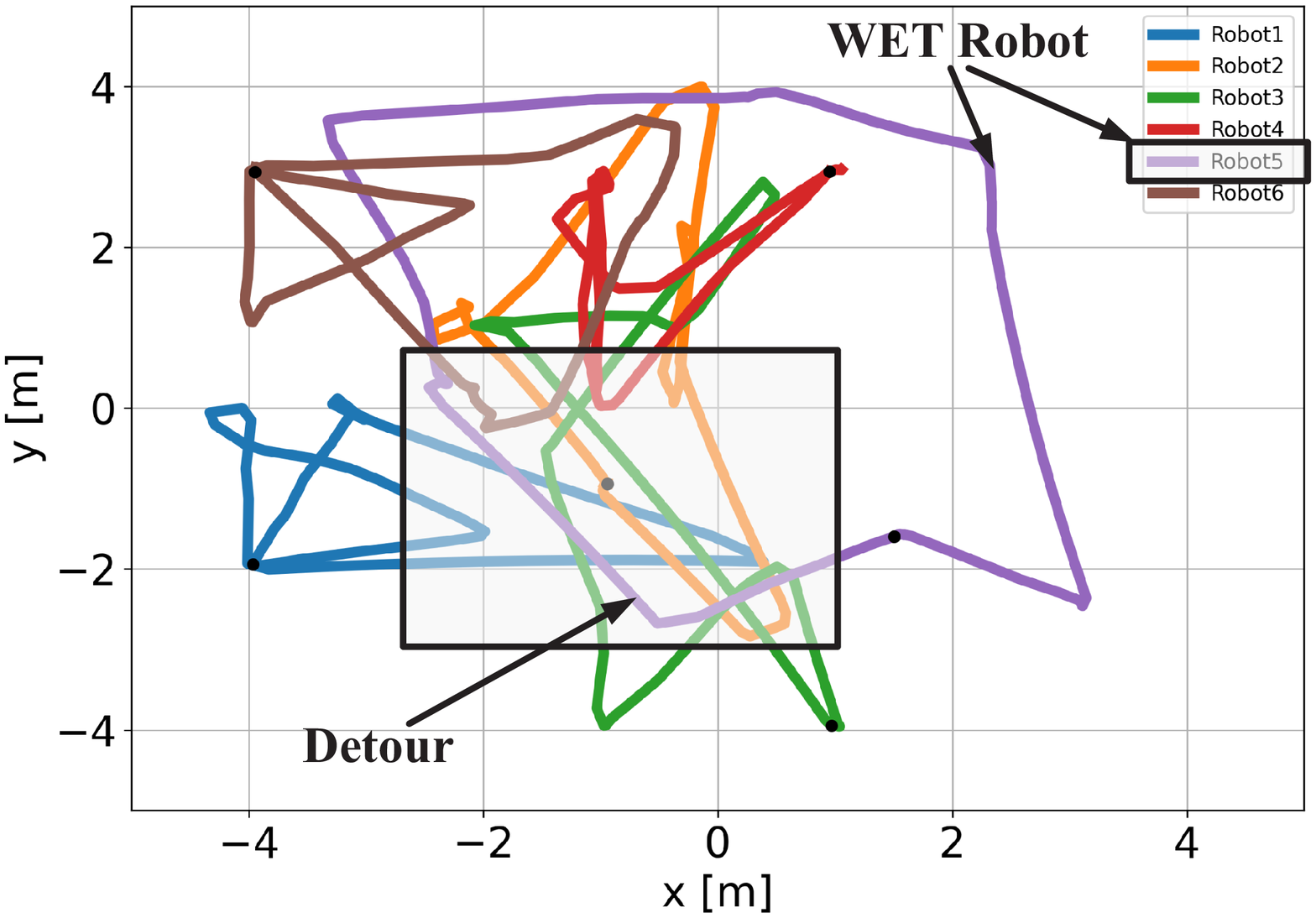}}
\subfigure[]{\includegraphics[width=0.6\textwidth]{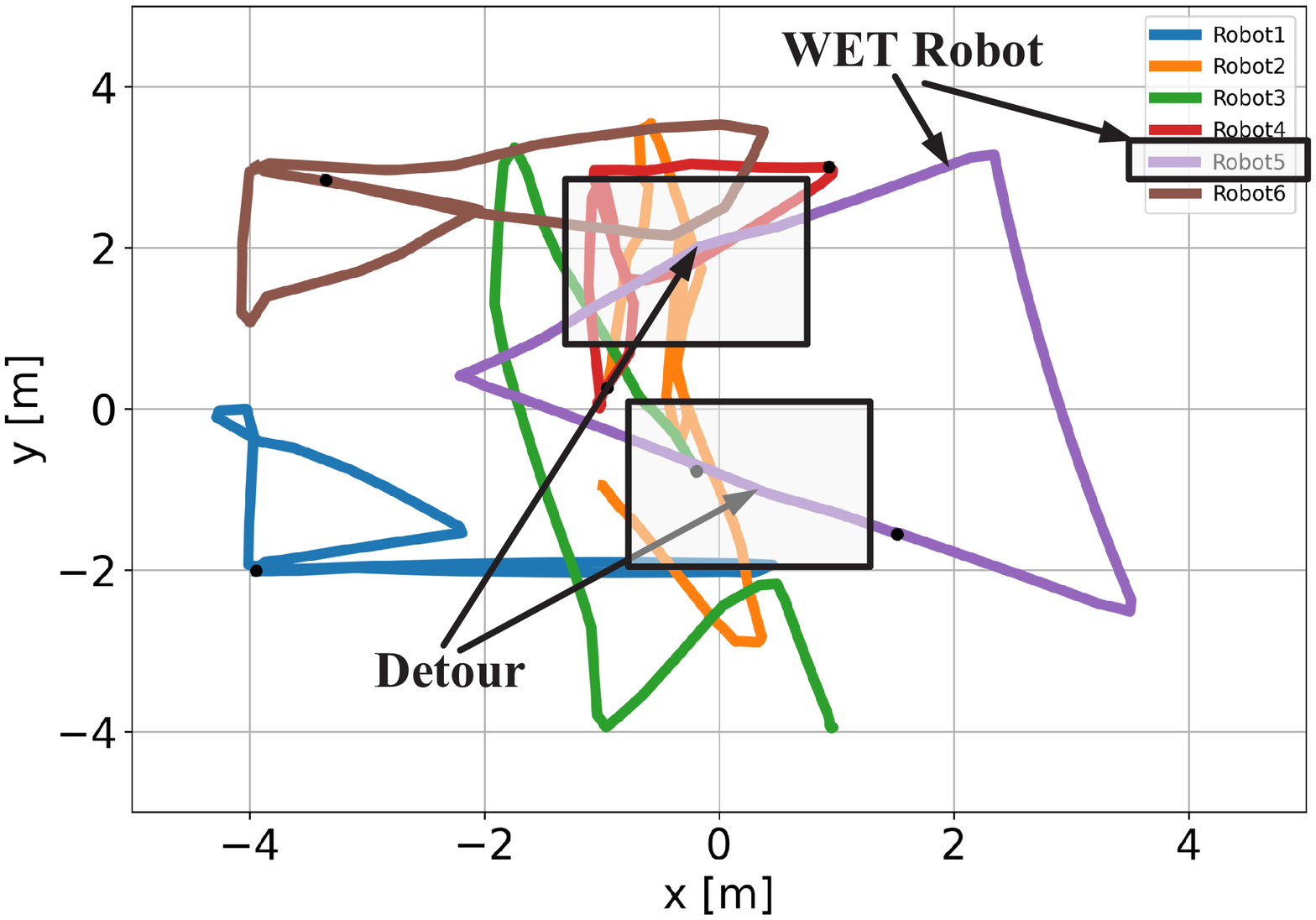}}
  \caption{Routes in Gazebo when 1 WET robot charges 20 EHs while avoiding collision with 5 noncooperative robots: a) before HIL re-optimization; b) after HIL re-optimization.}
\end{figure*}

\begin{figure*}[!t]
\centering
\includegraphics[width=0.98\textwidth]{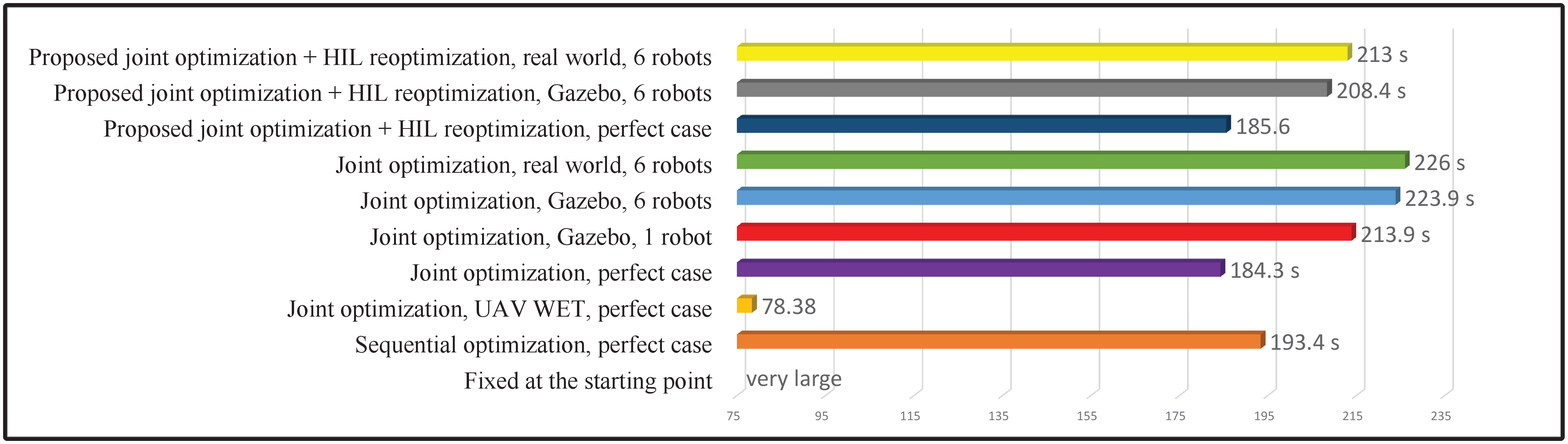}
\caption{Comparison of WET mission completion time for different schemes in the perfect case, the Gazebo, and the real world.}
\label{fig_sim}
\end{figure*}

With the above experimental data, we perform one-round HIL re-optimization.
In particular, we multiply the distance matrix for the motion time model by a factor of 1.5 and repeat the planning and validation.
The associated result is shown in Fig.~4b.
Compared with the result in Fig.~4a, the route evolves from a polygon to a triangle, whose route length is significantly smaller.
This implies that the re-optimized design automatically chooses to reduce the robot's movements and navigates the robot away from the ``dense traffic'' area (i.e., lower-left of Fig.~4).

Finally, to verify the effectiveness of the proposed framework, we consider the same task and compare the amount of WET mission completion time for different schemes.
The major findings are summarized below.
\begin{itemize}
\item[(i)] The scheme with a fixed energy transmitter cannot accomplish the charging task within a reasonable amount of time, since the transmit power of TX91503 is only 3\,Watts and remote EHs can only receive $\mu$W-level power, which is below the sensitivity threshold.

\item[(ii)]
If the robot does not perform anchor point selection and reaches all the anchors, although the charging time can be reduced for each IoT, the total time consumption increases.
This is because moving costs extra time and there is a trade-off between spending time on moving versus on charging in the robotic WET system.

\item[(iii)]
By jointly optimizing the anchor, route, time, and beam, the total amount of time is reduced to 184.3\,s.
Note that the task completion time (including flying and charging) for the UAV WET scheme is only 78.38\,s.
This is because the UAV does not need to avoid obstacles in the sky and it adopts an airspeed of 5\,m/s.
However, the UAV energy consumption is 78.38$\times$100=7838\,Joule, which is significantly larger than the robot energy consumption (i.e., 184.3$\times$9.3=1714\,Joule).

\item[(iv)]
With the same joint optimization scheme, the time spent in Gazebo is longer than that in the perfect case.
This is because in Gazebo, the robot needs to 1) gradually rotate itself at transition points, 2) take detours to avoid collision with EHs, and 3) accelerate/decelerate between two consecutive anchor points.

\item[(v)]
Increasing the number of robots from 1 to 6 in Gazebo leads to an obvious increment in the amount of time.
This implies that the traffic condition of the environment would have a non-negligible impact on the system performance.

\item[(vi)]
The task completion time in real world is slightly longer than that in Gazebo.
This is because Gazebo implementation is centralized while the real world implementation is distributed, which involves various uncertainties such as position inaccuracies and time asynchronization.

\item[(vii)]
Compared with joint optimization without HIL, the completion time of HIL joint optimization is slightly longer in the perfect case, but is significantly shorter in the real world experiment (or Gazebo simulator).
This implies that a good theoretical scheme may break down in practice due to the mismatch between models and environments and vice versa.
This also demonstrates the significance of the HIL framework, which achieves robust performance in real-world dynamic environments.
\end{itemize}

\section{Conclusion}

This paper reviewed existing robotic WET algorithms and proposed a new HIL system design framework.
By comparing the results in perfect case, Gazebo, and real world, it is found that the robotic charging time may significantly increase from ideal to practical environments due to robotic dynamics and collision avoidance.
It was shown that HIL re-optimization could improve the robustness of robotic WET against practical uncertainties.

When multiple robots serve as WET chargers, exploiting their interactions and collaborations can significantly enhance the system performance.
Moreover, if the environment is absolutely unknown, the robot needs to create a global map of the environment, which can be time-consuming.
Integrated sensing and communication (ISAC) is a promising technique to accelerate the map merging procedure, as ISAC allows each robot to build and share its local map simultaneously.

\appendices

\section{Mathematical Models}

\subsection{Robot Motion Time Model}

We use $\mathbf{v}=[v_1,\cdots,v_M]$ to denote the anchor point selection variable, where $v_m=1$ represents anchor $m$ being selected and $v_m=0$ otherwise.
The total number of anchor points is $M$.
We use $\mathbf{W}=[W_{1,1},\cdots,W_{M,M}]$ to denote the route planning variable, where $W_{m,j}=1$ means that the robot travels from anchor $m$ to $j$ and $W_{m,j}=0$ otherwise.
The distance matrix $\mathbf{D}$ is written as
$\mathbf{D}=[D_{1,1},\cdots,D_{1,M};\cdots;D_{M,1},\cdots,D_{M,M}]\in\mathbb{R}^{M\times M}_+$, with the element $D_{m,j}$ representing the distance from anchor $m$ to anchor $j$ ($D_{m,m}=0$ for any $m$).
Initially, $\mathbf{D}$ is calculated as distance over velocity, based on the locations of all anchor points.
Due to the necessity of obstacle avoidance, the actual distance would be longer than the initial values.
The matrix $\mathbf{D}$ is refined after each experimental validation.
We use $\alpha$ in m/s to denote the velocity of the robot.

The motion variables $\mathbf{v}$ and $\mathbf{W}$ satisfy the following constraints:
\begin{subequations}
\begin{align}
&v_{m}\in\{0,1\},~~\forall m, \label{vertex}
\\
&W_{m,j}\in\{0,1\},~~\forall m,j,~~W_{m,m}=0,~~\forall m, \label{edge}
\\
&
\sum_{j=1}^MW_{m,j}=v_m,~\sum_{j=1}^MW_{j,m}=v_m,~~\forall m, \label{P1c}
\\
&\lambda_m-\lambda_j+\left(\sum_{l=1}^Mv_l-1\right)W_{m,j}+\left(\sum_{l=1}^Mv_l-3\right)W_{j,m}
\nonumber\\
&
\leq \sum_{l=1}^Mv_l-2+J\left(2-v_m-v_j\right),\quad \forall m,j\geq 2,~m\neq j, \label{subtour1}
\\
&v_m\leq\lambda_m\leq\left(\sum_{l=1}^Mv_l-1\right)v_m,~~\forall m\geq 2, \label{subtour2}
\end{align}
\end{subequations}
where \eqref{vertex}--\eqref{edge} are the anchor and route selection constraints, \eqref{P1c}--\eqref{subtour2} guarantee that the route is a connected ring (where $J=10^3$ and $\{\lambda_m\}$ are slack variables to guarantee that no subtour is generated).

\subsection{Wireless Channel Model}

The wireless fading environment is modeled based on the 3GPP TR 38.901.
Let $h_{k,m}$ denote the channel between the anchor $m$ and the EH $k$.
According to the 3GPP InH-office pathloss model, the channel gain $|h_{k,m}|^2$ in dB is a function of the positions of the anchor $m$ and the EH $k$, which can be computed as
\begin{align}
&-10\,\mathrm{log}_{10}\left(|h_{k,m}|^2\right)=32.4+17.3\,\mathrm{log}_{10}\left(\|[x_{k}-a_m, y_{k}-b_m]\|\right)
+20\,\mathrm{log}_{10}\left(f_c\right),
\end{align}
where $(x_k, y_k)$ is the location of EH $k$, $(a_m, b_m)$ is the location of anchor $m$, and $f_c$ in Hz is the carrier frequency.

\subsection{Energy Beam Model}

\begin{figure}[!t]
 \centering
\includegraphics[width=60mm]{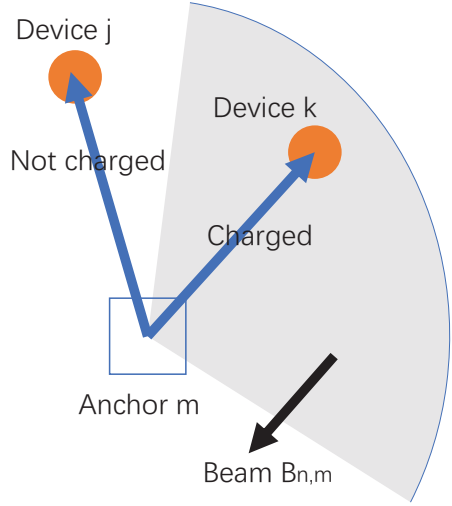}
  \caption{Illustration of the energy beam model.}
\end{figure}

The Powercast TX91503 adopts a single directional antenna with a beam width of $130$ degree.
Hence, the robot needs to rotate itself to generate different beam directions.
It is assumed that there are $N$ different beam directions in the codebook $\mathcal{C}=\{\mathcal{C}_{1},\cdots,\mathcal{C}_{N}\}$.
Each $\mathcal{C}_{n}$ defines a set of directions covered by the beam pattern $n$.
Now consider the case when the robot stops at anchor $m$ and adopts beam $\mathcal{B}_{n,m}\in\mathcal{C}$, as shown in Fig.~6.
If $[x_{k}-a_m, y_{k}-b_m]^T\in\mathcal{B}_{n,m}$, then the received power at EH $k$ is $P_T|h_{k,m}|^2$, where $P_T$ is the transmit power (including the antenna gain) and $h_{k,m}$ is the channel from anchor point $m$ to EH $k$.
Otherwise, the EH $k$ would receive no power.
Combining the two cases, the received RF power at EH $k$ is $P_T|h_{k,m}|^2\mathbb{I}_{[x_{k}-a_m, y_{k}-b_m]^T\in\mathcal{B}_{n,m}}$, where $\mathbb{I}$ is the indicator function.

\subsection{Energy Harvesting Model}

The sensitivity-based logistic model is given by:
\begin{align}
&\Upsilon(P_{\mathrm{in}})
=
\Bigg[\frac{P_{\mathrm{max}}}{\mathrm{e}^{-\tau P_{0}+\nu}}\left(\frac{1+\mathrm{e}^{-\tau P_{0}+\nu}}{1+\mathrm{e}^{-\tau P_{\mathrm{in}}+\nu}}-1\right)\Bigg]^+,
\end{align}
where $P_{\mathrm{in}}$ is the received radio frequency power, the parameter $P_{0}$ denotes the harvester's sensitivity threshold, and $P_{\mathrm{max}}$ refers to the maximum harvested power when the energy harvesting circuit is saturated.
The parameters $\tau$ and $\nu$ are used to capture the nonlinear dynamics of energy harvesting circuits.

Let $t_{n,m}$ denote the time duration when the robot stops at point $m$ and adopts beam $\mathcal{B}_{n,m}$.
The stopping time must be zero if the anchor is not visited, which yields
\begin{align}
&t_{n,m}\geq0,~~\forall n,m, \quad (1-v_m)\cdot t_{n,m}=0,~~\forall n,m. \label{notvisit}
\end{align}
The total energy harvested at EH $k$ satisfies
\begin{align}
\sum_{m=1}^M\sum_{n=1}^N
t_{n,m}\,\Upsilon\left(P_T|h_{k,m}|^2\mathbb{I}_{[x_{k}-a_m, y_{k}-b_m]^T\in\mathcal{B}_{n,m}}\right)\geq\gamma_k,~~\forall k,
\end{align}
where $\gamma_k$ in Joule is the energy harvesting requirement at EH $k$.

\section{Joint Optimization for Global Planning}

Let $\mathcal{G}_m$ denote the cluster of EHs associated with anchor $m$ after running the DBSCAN algorithm.
If the beam width of the transmitter is $360$ degree, the $m$-th K-Chebyshev anchor position is given by
\begin{align}
&(a_m^*,b_m^*)=\mathop{\mathrm{arg~min}}_{a_m,b_m}~\mathop{\mathrm{max}}_{k\in\mathcal{G}_m}\|[a_m-x_k,b_m-y_k]\|^2. \label{cheby}
\end{align}
If the beam width of the transmitter is smaller than $360$ degree, the $m$-th K-Chebyshev anchor position is modified into
\begin{align}
(a_m^*,b_m^*)=\mathop{\mathrm{arg~min}}_{a_m,b_m,\mathcal{B}}~&\mathop{\mathrm{max}}_{k\in\mathcal{G}_m}\|[x_k-a_m,y_k-b_m]\|^2,
\nonumber\\
\quad\quad\quad\quad\quad\quad\mathrm{s.t.}~~~~~&[x_{k}-a_m, y_{k}-b_m]^T\in\mathcal{B},\quad \forall k\in\mathcal{G}_m,
\nonumber\\
&\mathcal{B}\in\{\mathcal{C}_{1},\cdots,\mathcal{C}_{N}\}.
\label{cheby_bf}
\end{align}

Based on the mathematical models in Appendix A and the anchor positions in \eqref{cheby}--\eqref{cheby_bf}, the joint anchor point selection, route planning, and resource allocation problem is formulated as
\begin{subequations}
\begin{align}
\mathrm{P}:
\mathop{\mathrm{min.}}_{\substack{\mathbf{v},\mathbf{W},\{t_{n,m}\},\\\{\mathcal{B}_{n,m}\in\mathcal{C}\},\{\lambda_m\}}}
~~&\frac{1}{\alpha}\mathrm{Tr}(\mathbf{D}^{T}\mathbf{W})+\sum_{m=1}^M\sum_{n=1}^Nt_{n,m} \label{P1obj}
\\\mathrm{s.~t.}~~~~~&(1), (4), (5),
\end{align}
\end{subequations}
where \eqref{P1obj} is the operation time (including moving and charging).

\footnotesize

\vspace{0.5in}

\textbf{Shuai Wang} (s.wang@siat.ac.cn) received the Ph.D. degree from the University of Hong Kong. He is currently an Associate Professor with the Shenzhen Institute of Advanced Technology, Chinese Academy of Sciences.

\vspace{0.1in}

\textbf{Ruihua Han} (hanruihuaff@gmail.com) is a Ph.D. student with the Southern University of Science and Technology.

\vspace{0.1in}

\textbf{Yuncong Hong} (hongyc@mail.sustech.edu.cn) is a Ph.D. student with the Southern University of Science and Technology.

\vspace{0.1in}

\textbf{Qi Hao} (hao.q@sustech.edu.cn) received the Ph.D. degree from the Duke University. He is currently an Associate Professor with the Department of Computer Science and Engineering, Southern University of Science and Technology.

\vspace{0.1in}

\textbf{Miaowen Wen} (eemwwen@scut.edu.cn) received the Ph.D. degree from the Peking University. He is currently a Full Professor with the School of Electronic and Information Engineering, South China University of Technology.

\vspace{0.1in}

\textbf{Leila Musavian} (leila.musavian@essex.ac.uk) received her Ph.D. degree in Telecommunications from Kings College London, UK.
She is currently a Full Professor at the School of Computer Science and Electronic Engineering, University of Essex.

\vspace{0.1in}

\textbf{Shahid Mumtaz} (smumtaz@av.it.pt) received the Ph.D. degree from the University of Aveiro, Portugal. He is currently a Senior Research Scientist at the Instituto de Telecomunica\~{o}es.

\vspace{0.1in}

\textbf{Derrick~Wing~Kwan~Ng} (w.k.ng@unsw.edu.au) received his Ph.D. degree from The University of British Columbia.
He is currently an Associate Professor with the School of Electrical Engineering and Telecommunications, the University of New South Wales.
He is a Fellow of the IEEE.

\end{document}